\documentclass[times, review, 10pt]{elsarticle}

\usepackage{amssymb}
\usepackage{amsmath}
\usepackage{textcomp}
\usepackage{multicol}
\usepackage{multirow}
\usepackage{booktabs}
\usepackage{tabularx}
\usepackage{adjustbox}
\usepackage{caption}

\journal{Pattern Recognition}

\begin{document}

\begin{frontmatter}



\title{$\rm Med\text{-}R^{2}$: Perception and Reflection-driven Complex Reasoning for Medical Report Generation}


\author[label1]{Hao Wang} \author[label1]{Shuchang Ye} \author[label3]{Jinghao Lin} \author[label2]{Usman Naseem} \author[label1]{Jinman Kim\corref{cor1}}

\cortext[cor1]{Corresponding author: jinman.kim@sydney.edu.au}

\affiliation[label1]{organization={The School of Computer Science, The University of Sydney}, country={Australia}}
\affiliation[label2]{organization={The School of Computing, Macquarie University}, country={Australia}}
\affiliation[label3]{organization={Doubao Medical Group, ByteDance}, country={China}}

\begin{abstract}
Automated medical report generation (MRG) is increasingly used to reduce the burden of manual reporting and for decision support. Large vision-language models (LVLMs) hold great promise for automated MRG due to their fine-grained image-text alignment and advanced text-generation capabilities. Currently, state-of-the-art MRGs primarily focus on adapting pre-trained LVLMs with direct supervised fine-tuning (SFT), a fine-tuning strategy with medical image-report pairs. However, several factors limit the performance of these LVLMs. Firstly, direct SFT enables LVLMs to generate medical reports directly without an intermediate thinking process of pathological feature perception and diagnostic reasoning. This causes a potential failure to perceive pathological features and thus leads to misdiagnosis. Secondly, direct SFT lacks the incorporation of radiology-specific knowledge guidance, causing LVLMs to misinterpret perceived pathological features and make incorrect diagnoses. To address these gaps, we propose a novel fine-tuning strategy named $\rm Med\text{-}R^{2}$. We introduce a perception-driven long reasoning process that precedes report generation and incorporates radiology-specific knowledge as guidance. Additionally, to alleviate potential perceptual errors in complex reasoning, a reflection mechanism is introduced to refine the perception of pathological features and the generated report. Our experiments demonstrate that $\rm Med\text{-}R^{2}$ effectively enhances the capability of pathological features perception and diagnosis accuracy for MRG via fine-tuned LVLMs.
\end{abstract}



\begin{keyword}


Complex Reasoning \sep Medical Report Generation \sep Post-Training \sep Large Vision-language Model
\end{keyword}

\end{frontmatter}



\section{Introduction}
\label{intro}
Automated medical report generation (MRG) is an emerging technique aimed to relieve radiologists’ workload and improve diagnostic accuracy \cite{liu2023systematic,azad2024advances}. With the development of large vision–language models (LVLMs), these models have shown great potential for the MRG task owing to their image–text alignment and remarkable text generation capabilities \cite{hartsock2024vision}. Through large-scale pre-training on general datasets and medical-adaptive fine-tuning on medical datasets, LVLMs have demonstrated their ability to perform logical text generation while aligning the details of medical images with clinical terms \cite{sellergren2025medgemma,xu2025lingshu,liu2025vision}. Moreover, building on these advantages, LVLMs have shown the potential to conduct fine-grained interpretations of medical images and generate accurate medical reports with diagnostic evidence \cite{deng2025cross}.

LVLMs fine-tuned on medical datasets require further adaptation for MRG-related datasets to improve the performance of the MRG task. A common approach is via direct supervised fine-tuning (SFT), a fine-tuning strategy with medical-image pairs \cite{chen2024chexagent,boecking2022making}. LLaVA-Med \cite{li2023llava} proposed a rapid training framework for any medical downstream tasks (including the MRG task) that relies on direct SFT. Med-Flamingo \cite{moor2023med} was fine-tuned based on a general pre-trained model \cite{awadalla2023openflamingo} with an image-text dataset developed by the authors themselves. Based on the direct SFT strategy, multi-stage SFT strategies are widely explored for MRG adaptation. As an example, RadFM \cite{wu2023towards} was initialized with general pretrained MedLLaMA-13B and then fine-tuned with a custom-designed MedMD dataset to accumulate knowledge about medical-specific terminologies and images. Next, it was fine-tuned with the RadMD datasets, which consist of radiology image-text pairs. 

However, adapting LVLMs with these strategies still faces several limitations, especially in aspects of disease detection and diagnosis \cite{hartsock2024vision,messina2022survey,hou2025cross}. Firstly, the process of disease detection relies on accurate perceptions of pathological features. In current fine-tuning strategies, LVLMs are optimized to predict the next token, which corresponds to a latent linguistic embedding, without the assistance of explicit pathological or diagnostic tokens. Thus, perceptions of pathological features and the disease diagnosis must be completed within limited token lengths. However, foundation LVLMs used for adaptation only align medical images with pathological features at the visual level. They still struggle to align these pathological features to diagnosis in reports directly. Therefore, these strategies lead to insufficient perception of pathological features in fine-tuned LVLMs. Secondly, the interpretation from pathological features to disease diagnosis requires radiology-related knowledge for assistance. The process of current strategies also lacks the introduction of extra radiology-related knowledge for foundation LVLMs. Thus, this leads to potential misinterpretation of pathological features.

To address these limitations, we propose a new fine-tuning strategy, $\rm Med\text{-}R^{2}$, which incorporates a perception-driven long reasoning process before report generation and introduces radiology-specific knowledge to guide this process. In this process, perception is structured according to anatomical regions, with radiology-specific knowledge embedded as guidance for feature perception and diagnostic reasoning. A perception tree is further constructed based on clinical radiology datasets to provide reliable constraints for perception and knowledge recall. However, such a complex reasoning strategy may introduce additional perceptual errors. To mitigate this, we design a reflection mechanism that refines pathological perception and corrects report generation. Our contributions are as follows:

\begin{itemize}
    \item We propose $\rm Med\text{-}R^{2}$, a novel fine-tuning strategy that introduces a perception-driven long reasoning process to enhance LVLMs’ adaptation to the MRG task. The proposed complex reasoning process integrates radiology-specific knowledge and consists of anatomy-based pathological perception and fine-grained report generation under the guidance of a perception tree.
    \item We construct a perception tree based on real clinical data (MIMIC-CXR and IU-Xray). It consists of comprehensive pathological features and guides for the perception part of the complex reasoning process.
    \item We introduce a reflection mechanism applied to both the perception content generation and final report generation to mitigate potential perceptual errors, such as hallucinations, introduced by complex reasoning.
\end{itemize}

To validate the effectiveness of $\rm Med\text{-}R^{2}$, we fine-tune both generalist LVLMs and a medical pretrained LVLM (Qwen2.5VL-7B \cite{bai2025qwen2}, Llama3.2-Vision-11B \cite{dubey2024llama}, LLaVA-Med \cite{li2023llava}), and compare them with several state-of-the-art (SOTA) medical LVLMs (MedGemma \cite{sellergren2025medgemma}, MedVLM-R1 \cite{pan2025medvlm}, Lingshu \cite{xu2025lingshu}, etc.). Evaluations are conducted using standard natural language generation (NLG) metrics \cite{papineni2002bleu,banerjee2005meteor,lin2004rouge} and clinical efficacy (CE) metrics \cite{smit2020chexbert,liu2019clinically}, as well as case-based analyses to demonstrate the effectiveness of the complex reasoning and reflection mechanisms.

\section{Related Works}
\label{related works}
\subsection{Medical Report Generation}
Early work on MRG mainly focused on encoder–decoder architectures with structural improvements. Jing et al. \cite{jing2017automatic} introduced co-attention with a hierarchical LSTM and joint tag prediction to strengthen image–text alignment and handle long paragraph generation. HRGR-Agent \cite{yang2021writing} combined sentence-level template retrieval with free-form generation and used reinforcement learning to train the retrieval/generation policy, improving terminology coverage and linguistic flexibility. R2Gen \cite{chen2020generating} proposed a memory-driven Transformer with relational memory and conditional layer normalization, enabling the reuse of historical information across decoding steps and reducing cross-sentence inconsistencies. AlignTransformer \cite{you2021aligntransformer} predicted disease tags, aligned regions hierarchically, and decoded from multi-grained features to enhance evidence–text correspondence. Harzig et al. \cite{harzig1908addressing} introduced dual decoders for normal and abnormal sentences to address training imbalance, while Yuan et al. \cite{yuan2019automatic} fused multi-view and multi-scale features and jointly trained disease classification and report generation to improve abnormality coverage.

Beyond architectural improvements, incorporating medical knowledge has been another major research focus in MRG. PPKED \cite{liu2021exploring} combined posterior localization and prior knowledge extraction via multi-domain knowledge distillation to mitigate vision–language bias. KiUT \cite{huang2023kiut} employed a U-shaped cross-modal Transformer with skip connections, enhanced by symptom graphs and knowledge distillation modules to ensure clinical consistency. KERP \cite{li2019knowledge} encoded abnormality graphs, retrieved sentence templates, and paraphrased them to improve the traceability of findings. MedWriter \cite{yang2021writing} introduced dual retrieval at both report and sentence levels to provide prior context for decoding, achieving stronger clinical alignment on public datasets. Rule- and template-based systems \cite{pino2021clinically, kale2023replace} detected abnormalities and replaced or inserted sentences from predefined templates, providing stable reporting for known categories.

\subsection{Large Vision-language Models for Medical Report Generation}
LVLMs have recently been introduced for medical report generation (MRG) through both pre-training and fine-tuning strategies. Pre-trained medical LVLMs aim to build domain-specific cross-modal representations directly from medical data. For example, MedGemma \cite{sellergren2025medgemma} and Lingshu \cite{xu2025lingshu} are large medical-pretrained LVLMs trained on extensive radiology datasets. They learn detailed visual–textual correspondence and are further refined for MRG, improving diagnostic reliability without requiring heavy instruction tuning. Similarly, Flamingo-CXR \cite{tanno2023consensus}, based on the generalist model Flamingo \cite{alayrac2022flamingo}, integrates a pre-trained visual encoder and language decoder and is fine-tuned on medical image–text pairs for report generation. In a study with board-certified radiologists \cite{tanno2025collaboration}, Flamingo-CXR’s reports were rated comparable to clinician-written reports in most cases.

To alleviate the gap mentioned above, the SFT strategy is employed to fine-tune pre-trained LVLMs to adapt to the MRG task. LLaVA-Med \cite{li2023llava}, an extension of the LLaVA foundation model, was trained on large-scale medical datasets and achieved improved responses to medical-related questions compared with LLaVA. Its two-phase fine-tuning includes domain alignment on biomedical figures and captions, followed by instruction tuning on GPT-4–generated QA dialogues, demonstrating that foundation models can be efficiently adapted using medical VQA data. RadFM \cite{wu2023towards} fine-tuning on various radiology-related datasets based on the ViT encoder and MedLLaMA-13B decoder structure.

\subsection{Reasoning in Large Vision-language Models}
Chain of thought (CoT) in LVLMs extends step-by-step prompting from language models to guide intermediate reasoning before answers, improving performance on complex visual tasks. Multimodal-CoT \cite{zhang2023multimodal} uses a two-stage framework that first generates a detailed textual rationale integrating visual and textual cues, then infers the answer. Similarly, Lu et al. \cite{liu2023mitigating} showed that prompting models to explain their answers to science diagram questions improves accuracy via multimodal thought chains. Zheng et al. \cite{zheng2023ddcot} proposed Duty-Distinct CoT, which separates visual description from logical reasoning and incorporates negative cues to enhance reasoning accuracy. Wang et al. \cite{wang2022towards} introduced self-consistency for LVLMs by sampling multiple reasoning chains and selecting the majority answer to improve reliability. Overall, these methods demonstrate that structured reasoning before answer generation enhances LVLM reasoning on complex visual tasks \cite{cheng2024vision}, but they rely heavily on external prompts and the underlying reasoning ability of foundation models, and errors in intermediate steps can still propagate through the entire chain.

To address these limitations, recent research has focused on making the reasoning process more interpretable and grounded in visual and domain-specific content. Visual CoT \cite{shao2024visual} introduced a multi-turn reasoning pipeline that iteratively focuses on different image regions, generating intermediate visual reasoning steps. A visual CoT dataset of 438k QA pairs with annotated bounding boxes and textual reasoning steps was curated to train and evaluate this behavior. Furthermore, GPT-o1 introduced a long CoT strategy that enables complex and fine-grained reasoning. Instead of relying solely on prompting, GPT-o1 was trained via RL and SFT to generate and refine long CoT internally. In the medical domain, MedVLM-R1 \cite{pan2025medvlm} extended this line of work by applying RL-based reasoning to medical VQA tasks, effectively enhancing domain-specific reasoning and diagnostic interpretation. However, the RL-only strategy relies heavily on model self-learning and lacks external guidance, leading to limited generalization in complex medical scenarios.

\begin{figure*}[!t]
\centering
\includegraphics[width=\linewidth]{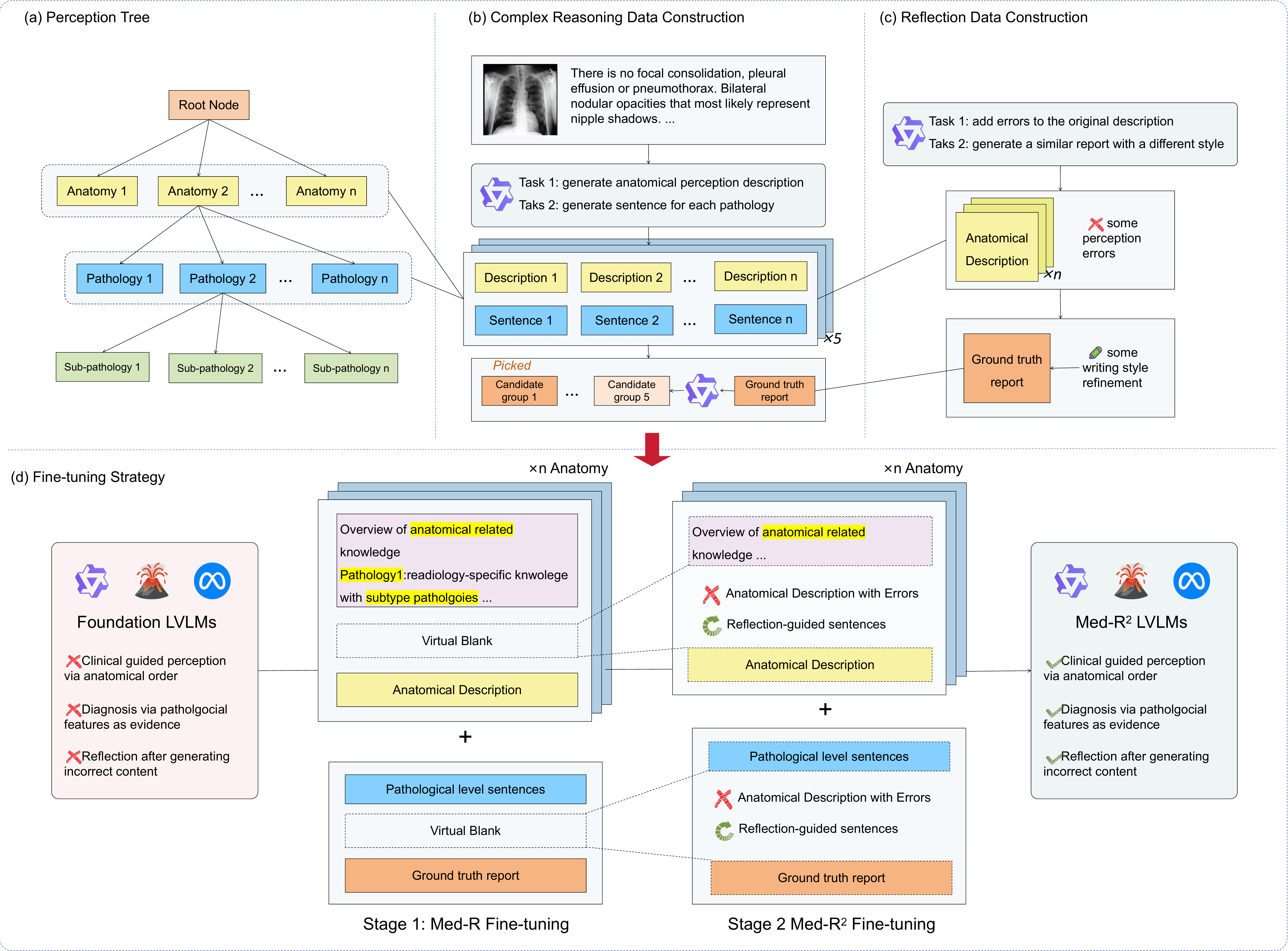}
\caption{Overview of the proposed $\mathrm{Med\text{-}R^{2}}$ framework. (a) A perception tree organizes anatomical structures, pathologies, and sub-pathologies to enforce clinically ordered perception. (b) Complex reasoning data construction generates anatomy-level descriptions and pathology-level sentences, from which high-quality candidates are selected against ground-truth reports. (c) Reflection data construction injects perception errors into anatomical descriptions and refines them via reflection to improve report quality and style. (d) A two-stage fine-tuning strategy ($\mathrm{Med\text{-}R}$ and $\mathrm{Med\text{-}R^{2}}$) progressively enhances clinically guided perception, pathology-based diagnosis, and reflection after incorrect generation.}
\label{fig:flow}
\end{figure*}

\section{Methodology}
The $\rm Med\text{-}R^{2}$ is a fine-tuning strategy designed to guide LVLMs toward a pathology-perception-centered reasoning process before producing final reports, enabling revision of perception content to improve diagnostic accuracy. It adopts a staged pipeline in which the model gradually learns to observe, reason, and refine rather than generate a report in one step. First, a hierarchical perception structure is introduced to organize anatomical regions and related pathologies as the foundation for reasoning. The model is then trained to generate intermediate reasoning content and fine-grained diagnostic statements based on this structure. Finally, a reflection mechanism is applied to review and revise perception outputs and report drafts, alleviating perceptual hallucination and improving accuracy. Fig. \ref{fig:flow} shows the three key steps: (i) defining and constructing the perception tree; (ii) fine-tuning LVLMs with medical images, reasoning processes, and reports; and (iii) fine-tuning LVLMs with reasoning augmented by reflection.

\begin{figure*}[]
\centering
\includegraphics[width=\linewidth]{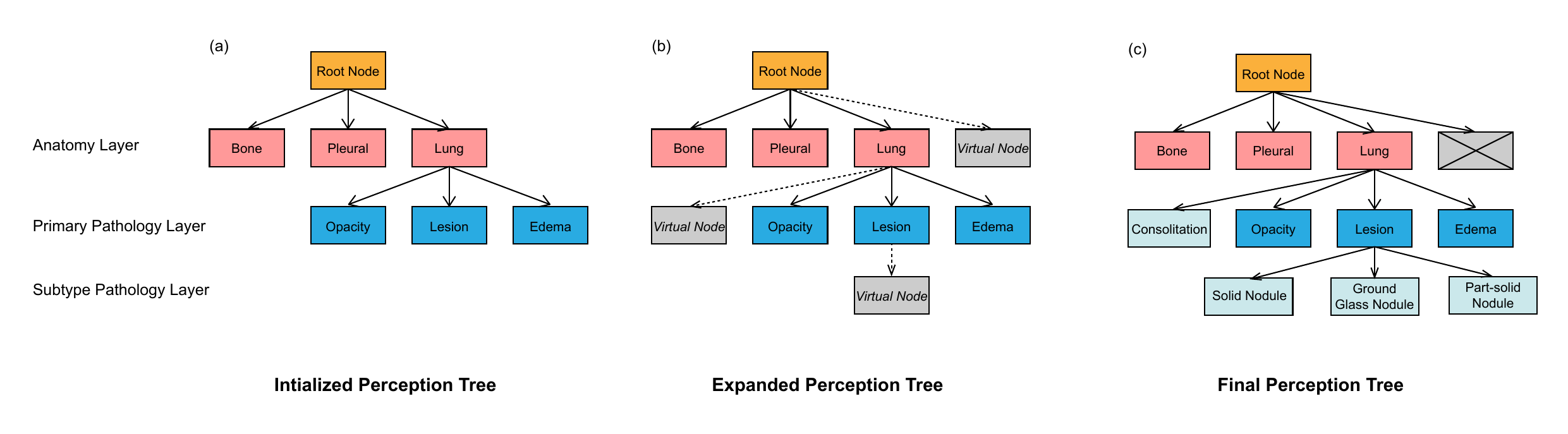}
\caption{Perception tree construction process: (a) initialization with prior knowledge graph; (b) incorporation of virtual nodes and classification of report sentences; (c) expansion and pruning to derive the final perception tree.}
\label{fig:pt}
\vspace{-\baselineskip}
\end{figure*}

\subsection{Perception Tree Construction}
\textbf{Perception tree definition}. The perception tree consists of three hierarchical layers and two node types: anatomy nodes and pathology nodes. As shown in Fig. \ref{fig:pt}, the first layer represents major anatomical structures in chest X-rays, such as lungs, heart, and ribs. The second and third layers are made up of pathology nodes, which include the primary diseases affecting each anatomical structure (primary pathology nodes) as well as their subtypes, categorized by distinct pathological features (subtype pathology nodes).

\textbf{Perception tree initialization}: The perception tree is initialized by pruning and adapting the chest abnormality graph proposed by Zhang et al. \cite{zhang2020radiology}. Specifically, anatomical nodes are initialized using the anatomical structure nodes in the graph, while the primary pathology nodes are initialized using the disease nodes. For further expansion, the subtype pathology nodes are set to blank as virtual nodes. In the primary pathology layer and the anatomical structure layer, virtual nodes are also added for further expansion, separately. 

\textbf{Building the Perception Tree}: To expand the initial perception tree following clinically realistic guidelines, we randomly extracted 7k medical reports from the IU-Xray and MIMIC-CXR datasets as references. These reports were divided into sentences and assigned to different anatomy nodes by an LLM-based agent based on their semantic content. Sentences that did not fit existing anatomy nodes were assigned to the virtual anatomy node. For each anatomical group, another classifier associated sentences with pathology nodes and placed unmatched cases under virtual primary or subtype pathology nodes.

Building on this, the subsequent step is to reconstruct the subtrees associated with the virtual nodes. For each virtual node, we applied the K-means clustering algorithm to group sentences which embedded by CheXbert into 15 clusters. Each cluster was summarized into a subtype pathology under the corresponding pathology node. For virtual subtype pathology nodes, the newly derived nodes were added beneath the original pathology nodes. For virtual primary pathology nodes, the newly derived nodes were summarized by an LLM agent and attached to the relevant anatomy node. The same expansion strategy is applied to the virtual anatomy nodes as well.

\textbf{Pruning the perception tree}: After expanding the perception tree, pruning and merging were applied to remove redundancy. We systematically review each added subtree, assess its semantic consistency and clinical relevance, and eliminate nodes with ambiguous, redundant, or nonsensical meanings. Finally, the perception tree was obtained and served as the foundation for guiding complex reasoning processes.

\subsection{Complex Reasoning}
\textbf{Reasoning path design}: The complex reasoning process is designed with three key components: radiology-specific knowledge recall, pathological feature perception within anatomical structures, and fine-grained report generation. It is divided into the perception-tree guided perception stage and the fine-grained report generation stage.

\textit{Pathology perception within anatomical structures}. The pipeline of the perception stage follows the clinical diagnostic process \cite{bate2012clinical}, which ensures the effectiveness and reliability of the perception content. Guided by the nodes in the anatomical layer of the perception tree, the perception contents are generated according to the order of the given anatomical structures. However, when perception contents are directly generated for each anatomical structure, irrelevant pathological features may be incorporated, leading to the neglect of important pathological evidence during diagnosis. Therefore, a paragraph of radiology-specific knowledge is added before the perceptual process within each anatomical structure. This knowledge component is designed to guide the extraction of effective pathological features and the diagnostic reasoning based on corresponding pathological evidence. The knowledge is constructed upon the second and third layers of the perception tree. Specifically, it first enumerates the primary pathologies and provides diagnostic approaches grounded in their pathological evidence. Furthermore, each diagnostic approach is further refined into multiple sub-methods corresponding to different pathological subtypes.

\textit{Fine-grained report generation}. Although various pathological features are extracted from images, the generated report has a limited length to incorporate these features. Therefore, direct report generation referring to the previous perception contents may cause the omission of important pathological diagnoses or the abundant description of normal anatomical structures. To alleviate this limitation, sentence-level reports for primary pathology are generated before the final version report. This previous generation walks over all the primary pathology nodes in the second layer of the perception tree and generates one description sentence for each node. The final version report fuses these previous sentence-level reports and filters the abundant content as the final answer.

\textbf{Training data construction}: To enable LVLMs to learn the process of complex reasoning effectively, the SFT strategy is adopted for fine-tuning. We design a pipeline in which two agents collaborate within a unified framework to construct the complex reasoning process for each medical report. In addition, a general reasoning template is designed, leaving placeholders for each organ’s perception, fine-grained report sentences, and the final report. The input and output data for the first-stage fine-tuning are defined as follows: (1) medical images together with a prompt template, and (2) responses that combine complex reasoning data with the corresponding medical report.

\textit{Perception construction within anatomical structures}. To minimize errors in the generated content, the generation agent is provided with both the medical image and the corresponding report and tasked with producing perception content for each anatomical structure. This process is repeated 5 times, yielding 5 candidate outputs. To identify the most accurate perception content among these candidates, a verifier agent is employed. The verifier agent generates a report based solely on the perception content, without access to the medical image or the original report. Subsequently, similarity is computed between the verifier’s generated report and the ground truth report, with both representations embedded using CheXbert. The perception content with the highest similarity score is selected as the final SFT data.

\textit{Fine-grained report sentence construction}. Similar to organ perception construction, the generation agent is provided with the medical image–report pair and tasked with producing descriptive sentences for each primary pathology represented in the perception tree. For each pathology, the agent must determine whether the disease is present and generate a corresponding report-style sentence. In total, 5 groups of fine-grained sentences are generated as candidates. To verify the accuracy of these groups, a preview medical report is constructed by selecting and merging the fine-grained pathology sentences alone. Following the verification procedure used for organ perception data, both the preview medical report and the ground-truth report are embedded using CheXbert, and their similarity is computed. The group of fine-grained sentences with the highest similarity score is then selected as the most accurate.

\subsection{Reflection Mechanism}
\textbf{Reflection component design}. The process of complex reasoning introduces relative medical knowledge and perception contents, both guided by the perception tree. However, extra tokens of the perceptual process increase the potential of perception errors. Also, the fine-grained report generation introduces the potential of inappropriate merging. Therefore, to alleviate these potential errors, reflection components are inserted in two parts of complex reasoning: 1) perception contents generation, 2) final reports generation.

\textit{Perception content reflection}. After generating perception content for each organ, a guided sentence is introduced to evaluate potential errors and trigger regeneration when necessary. If the perception content is judged to be incorrect, a revised version is then generated. This reflection component helps mitigate potential diagnostic errors arising from flawed perception.

\textit{Fine-grained report generation reflection}. Similarly, a guided reflection sentence is embedded after the final report generation. If inconsistencies or errors are detected, the final report is forced to be regenerated. This rewriting process can reorganize sentence order and adjust writing style, making the report more consistent with clinical conventions.

\textbf{Reflection data construction}. To embed reflection components into the complex reasoning process, LVLMs are fine-tuned on reflection-embedded reasoning data with the SFT strategy as the second-stage training. We revise the complex reasoning template and add two blank placeholders respectively after the perception content and the final report. Also, a pipeline with two-agent collaboration is designed to generate reflection data automatically. 

\textit{Perceptual reflection data}. In constructing complex reasoning data, descriptive content is generated for each organ. To support reflection, the generation agent, provided with the ground-truth report, is required to modify the descriptions by introducing certain perception errors. The incorrect description, the correct description, and the reflection-guided sentence are then combined to form the reflection process and inserted into the designated placeholder.

\textit{Final report reflection data}. To ensure the accuracy of the preview version of the final report, the group of disease descriptions generated during the complex reasoning process is provided to the generation agent to produce a preview version of the final report. The preview report, the ground-truth report, and the reflection-guided sentence are then combined and inserted into the blank placeholder of the revised template.

\section{Experiments Setup}
\subsection{Datasets}
Experiments were conducted on two public X-ray image-text pair datasets. IU-Xray \cite{demner2016preparing} with 7,470 pairs and MIMIC-CXR \cite{johnson2019mimic} with 371,920 pairs. For the IU-Xray dataset, the training, validation, and testing split ratio of 7:1:2 was adopted following previous similar studies \cite{chen2022cross,chen2020generating,wang2022cross}. For the MIMIC-CXR dataset, the official data split was used.

\subsection{Evaluation Metrics}
Natural language generation (NLG) metrics and clinical efficacy (CE) metrics were employed, consistent with prior works on LVLMs. The NLG metrics included BLEU-1 \cite{papineni2002bleu}, BLEU-4 \cite{papineni2002bleu}, METEOR \cite{banerjee2005meteor}, and ROUGE-L \cite{lin2004rouge}, which were used to assess linguistic quality. The CE metrics \cite{liu2019clinically}, including precision, recall, and F-score, measure diagnostic accuracy. Since CE metrics were calculated based on classification vectors derived from the reports, CheXbert, an embedding model, was used to obtain these classification vectors.

\subsection{Baselines}
The fine-tuning strategy is implemented with two stages. Firstly, foundation LVLMs were fine-tuned with complex reasoning data via the SFT strategy. Sequentially, they were further fine-tuned on complex reasoning data that incorporate reflective components via the SFT strategy. The first stage of the fine-tuning strategy was named $\rm Med\text{-}R$. Also, the whole fine-tuning strategy was named $\rm Med\text{-}R^{2}$. To ensure that improvements were not coupled to a single model architecture, we applied the strategy across multiple LVLM backbones. We compared $\rm LlamaMed\text{-}R$ and $\rm LlamaMed\text{-}R^{2}$ (Llama3.2-Vision-11B fine-tuned with these two strategies separately) with current SOTA medical LVLMs (LLaVA-Med \cite{li2023llava}, RadFM \cite{wu2023towards}, MedVLM-R1 \cite{pan2025medvlm}, MedGemma \cite{sellergren2025medgemma}, Lingshu \cite{xu2025lingshu}). Furthermore, to assess the effectiveness of our approach, three pre-trained foundation models were fine-tuned with $\rm Med\text{-}R$ and $\rm Med\text{-}R^{2}$: Qwen2.5VL-7B, Llama3.2-Vision-11B, and LLaVA-Med (7B), and compared their performance with the direct SFT strategy.

\subsection{Implementation Detail}
For the automated construction pipeline of complex reasoning and reflection data, the generation agent and verifier were implemented using Qwen2.5VL-72B and Qwen2.5-72B, respectively. A total of 18k samples were randomly selected from the base datasets (4k from IU-Xray and 14k from MIMIC-CXR) to construct the complex reasoning and reflection data. The fine-tuning stages were conducted within the LLaMA-Factory framework using the LoRA strategy.

\begin{table*}[!t]
\centering
\captionsetup{skip=8pt}
\caption{Performance comparisons with LVLMs designed for report generation. ($\mathrm{LlamaMed\text{-}R}$ and $\mathrm{LlamaMed\text{-}R^{2}}$ indicates the Llama fine-tuned with the $\mathrm{Med\text{-}R}$ strategy and the $\mathrm{Med\text{-}R^{2}}$ strategy, respectively.)}
\label{tab:compare}
\begin{adjustbox}{center}
\fontsize{8pt}{8pt}\selectfont
\begin{tabularx}{1.05\textwidth}{l|l|*{4}{>{\centering\arraybackslash}X}|*{3}{>{\centering\arraybackslash}X}}
    \toprule
    Datasets & Methods & BL-1 & BL-4 & MTR & RG-L & P & R & F \\
    \midrule
    \multirow{5}{*}{IU-Xray}
    & LLaVA-Med & 0.066 & 0.026 & 0.070 & 0.132 & - & - & - \\
    & RadFM & 0.059 & 0.022 & 0.066 & 0.087 & - & - & - \\
    & MedVLM-R1 & 0.079 & 0.006 & 0.132 & 0.107 & - & - & - \\
    & MedGemma-4B & 0.199 & 0.029 & 0.263 & 0.235 & - & - & - \\
    & Lingshu & 0.252 & 0.061 & \textbf{0.268} & 0.297 & - & - & - \\
    \cmidrule{2-9}
    & $\rm LlamaMed\text{-}R$ & \underline{0.283} & \textbf{0.178} & 0.198 & \underline{0.298} & - & - & - \\
    & $\rm LlamaMed\text{-}R^{2}$ & \textbf{0.303} & \underline{0.169} & \underline{0.259} & \textbf{0.305} & - & - & - \\
    \midrule
    \multirow{5}{*}{MIMIC-CXR}
    & LLaVA-Med & 0.181 & 0.008 & 0.084 & 0.120 & 0.111 & 0.196 & 0.142 \\
    & RadFM & 0.040 & 0.006 & 0.054 & 0.090 & 0.076 & 0.092 & 0.042 \\
    & MedVLM-R1 & 0.130 & 0.007 & 0.129 & 0.132 & 0.082 & 0.110 & 0.065 \\
    & MedGemma-4B &  0.195 & 0.017 & 0.169 & 0.175 & 0.192 & 0.210 & 0.210 \\
    & Lingshu & 0.162 & 0.029 & 0.169 & 0.205 & 0.166 & 0.243 & 0.185 \\
    \cmidrule{2-9}
    & $\rm LlamaMed\text{-}R$ & \underline{0.269} & \underline{0.093} & \textbf{0.214} & \underline{0.211} & \underline{0.300} & \textbf{0.371} & \underline{0.330} \\
    & $\rm LlamaMed\text{-}R^{2}$ & \textbf{0.334} & \textbf{0.120} & \underline{0.195} & \textbf{0.273} & \textbf{0.370} & \underline{0.339} & \textbf{0.354} \\
    \bottomrule
\end{tabularx}
\end{adjustbox}
\end{table*}

\section{Results}
\subsection{Comparison of performance with medical LVLMs}
Table \ref{tab:compare} compares the performance of state-of-the-art medical LVLMs with $\rm LlamaMed\text{-}R$ and $\rm LlamaMed\text{-}R^{2}$ on the NLG and CE metrics. On the IU-Xray dataset, the $\rm LlamaMed\text{-}R^{2}$ achieved the best performance on BLEU-1 (0.303) and ROUGE-L (0.305), while the $\rm LlamaMed\text{-}R$ achieved the highest BLEU-4 score (0.178). On the MIMIC-CXR dataset, the $\rm LlamaMed\text{-}R$ outperformed the SOTA medical LVLMs on both the NLG metrics (BLEU-1: 0.269, BLEU-4: 0.093, METEOR: 0.217, ROUGE-L: 0.211) and the CE metrics (Precision: 0.300, Recall: 0.371, F-score: 0.330), and this was further improved by $\rm LlamaMed\text{-}R^{2}$ (BLEU-1: 0.334, BLEU-4: 0.120, ROUGE-L: 0.273) and the CE metrics (Precision: 0.370, F-score: 0.354).

\begin{table*}[!t]
\centering
\captionsetup{skip=8pt}
\caption{Evaluation results on Qwen2.5VL-7B, Llama3.2-Vision-11B and LLaVA-Med with different strategies (direct SFT, $\rm Med\text{-}R$, $\rm Med\text{-}R^{2}$). The highest performance is indicated in bold, while the second highest is marked with an underline.}
\label{tab:ablation}
\begin{adjustbox}{center}
\fontsize{8pt}{8pt}\selectfont
\begin{tabularx}{1.05\textwidth}{l|l|*{4}{>{\centering\arraybackslash}X}|*{3}{>{\centering\arraybackslash}X}}
    \toprule
    Datasets & Methods & BL-1 & BL-4 & MTR & RG-L & P & R & F \\
    \midrule
    \multirow{12}{*}{IU-Xray}
    & Qwen2.5VL-7B & 0.060 & 0.007 & 0.090 & 0.135 & - & - & - \\
    & +direct SFT & 0.210 & 0.067 & 0.223 & 0.231 & - & - & - \\
    & +$\rm Med\text{-}R$ & \underline{0.325} & \underline{0.165} & \textbf{0.247} & \underline{0.294} & - & - & - \\
    & +$\rm Med\text{-}R^{2}$ & \textbf{0.331} & \textbf{0.178} & \underline{0.226} & \textbf{0.308} & - & - & - \\
    \cmidrule{2-9}
    & Llama3.2-Vision-11B & 0.046 & 0.011 & 0.085 & 0.131 & - & - & - \\
    & +direct SFT & 0.179 & 0.057 & 0.198 & 0.218 & - & - & - \\
    & +$\rm Med\text{-}R$ & \underline{0.283} & \textbf{0.178} & \underline{0.210} & \underline{0.298} & - & - & - \\
    & +$\rm Med\text{-}R^{2}$ & \textbf{0.303} & \underline{0.169} & \textbf{0.259} & \textbf{0.305} & - & - & - \\
    \cmidrule{2-9}
    & LLaVA-Med & 0.066 & 0.026 & 0.070 & 0.132 & - & - & - \\
    & +direct SFT & 0.196 & 0.068 & 0.220 & 0.234 & - & - & - \\
    & +$\rm Med\text{-}R$ & \underline{0.290} & \underline{0.187} & \textbf{0.285} & \underline{0.313} & - & - & - \\
    & +$\rm Med\text{-}R^{2}$ & \textbf{0.347} & \textbf{0.199} & \underline{0.267} & \textbf{0.317} & - & - & - \\
    \midrule
    \multirow{12}{*}{MIMIC-CXR}
    & Qwen2.5VL-7B & 0.119 & 0.010 & 0.136 & 0.128 & 0.132 & 0.125 & 0.122 \\
    & +direct SFT & 0.165 & 0.036 & 0.139 & 0.199 & 0.237 & 0.295 & 0.259 \\
    & +$\rm Med\text{-}R$ & \underline{0.315} & \underline{0.075} & \underline{0.183} & \underline{0.256} & \underline{0.291} & \underline{0.324} & \underline{0.307} \\
    & +$\rm Med\text{-}R^{2}$ & \textbf{0.330} & \textbf{0.094} & \textbf{0.186} & \textbf{0.286} & \textbf{0.325} & \textbf{0.346} & \textbf{0.328} \\
    \cmidrule{2-9}
    & Llama3.2-Vision-11B & 0.102 & 0.007 & 0.071 & 0.113 & 0.153 & 0.159 & 0.147 \\
    & +direct SFT & 0.171 & 0.040 & 0.144 & 0.205 & 0.195 & 0.303 & 0.238 \\
    & +$\rm Med\text{-}R$ & \underline{0.269} & \underline{0.093} & \textbf{0.214} & \underline{0.211} & \underline{0.300} & \textbf{0.371} & \underline{0.330} \\
    & +$\rm Med\text{-}R^{2}$ & \textbf{0.334} & \textbf{0.120} & \underline{0.195} & \textbf{0.273} & \textbf{0.370} & \underline{0.339} & \textbf{0.354} \\
    \cmidrule{2-9}
    & LLaVA-Med & 0.181 & 0.008 & 0.084 & 0.120 & 0.111 & 0.196 & 0.142 \\
    & +direct SFT & 0.157 & 0.034 & 0.106 & 0.179 & 0.192 & 0.281 & 0.229 \\
    & +$\rm Med\text{-}R$ & \underline{0.235} & \underline{0.048} & \underline{0.162} & \underline{0.225} & \underline{0.279} & \underline{0.230} & \underline{0.252} \\
    & +$\rm Med\text{-}R^{2}$ & \textbf{0.287} & \textbf{0.051} & \textbf{0.183} & \textbf{0.246} & \textbf{0.283} & \textbf{0.294} & \textbf{0.288} \\
    \bottomrule
\end{tabularx}
\end{adjustbox}
\end{table*}

\subsection{Ablation Results of Reasoning and Reflection}
Table \ref{tab:ablation} presents the performance of Qwen2.5VL-7B, Llama3.2-Vision-11B, and LLaVA-Med on IU-Xray and MIMIC-CXR under the four different fine-tuning strategies: no fine-tuning, direct SFT, $\rm Med\text{-}R$, and $\rm Med\text{-}R^{2}$. These three foundation LVLMs fine-tuned with $\rm Med\text{-}R$ are named as $\rm QwenMed\text{-}R$, $\rm LlamaMed\text{-}R$, and $\rm LLaVAMed\text{-}R$ separately. Similarly, models fine-tuned with $\rm Med\text{-}R^{2}$ are named as $\rm QwenMed\text{-}R^{2}$, $\rm LlamaMed\text{-}R^{2}$, and $\rm LLaVAMed\text{-}R^{2}$ separately. Sequentially, Llama3.2-Vision-11B and its fine-tuned models are used as an example to illustrate the performance of the NLG metrics and CE metrics.

On the IU-Xray dataset, the foundation model without fine-tuning exhibited limited performance of the NLG metrics (0.046 of BLEU-1, 0.011 of BLEU-4, 0.085 of METEOR, and 0.131 of ROUGE-L). After fine-tuning with the direct SFT strategy, the performance improved to 0.179 of BLEU-1, 0.057 of BLEU-4, 0.210 of METEOR, and 0.218 of ROUGE-L. However, $\rm Med\text{-}R$ caused an overall promotion higher than the direct SFT. Compared with the direct SFT, $\rm LlamaMed\text{-}R$ showed consistent gains across all NLG metrics, particularly in BLEU-4 (+0.121) and BLEU-1 (+0.104), highlighting the benefit of incorporating complex reasoning during fine-tuning. Furthermore, $\rm Med\text{-}R^{2}$, which was further fine-tuned based on $\rm Med\text{-}R$, achieved additional improvements in the NLG metrics, especially in BLEU-1 and ROUGE-L. Compared with $\rm LlamaMed\text{-}R$, $\rm LlamaMed\text{-}R^{2}$ achieved increases of 0.020 in BLEU-1, demonstrating that the incorporation of the reflection mechanism further enhanced the model’s performance.

On the MIMIC-CXR dataset, the foundation model without fine-tuning demonstrated relatively limited performance on both the NLG and CE metrics (BLEU-1: 0.102, BLEU-4: 0.007, METEOR: 0.071, ROUGE-L: 0.113; Precision: 0.153, Recall: 0.159, and F-score: 0.147). After applying the direct SFT strategy, the performance improved to 0.171 of BLEU-1, 0.040 of BLEU-4, 0.144 of METEOR, and 0.205 of ROUGE-L, along with a substantial rise in the CE metrics (Precision: 0.195, Recall: 0.303, and F-score: 0.238). The $\rm LlamaMed\text{-}R$ model outperformed the direct SFT across both NLG and CE metrics, achieving 0.269 of BLEU-1, 0.093 of BLEU-4, 0.214 of METEOR, and 0.211 of ROUGE-L, with corresponding Precision, Recall, and F-score values of 0.300, 0.371, and 0.330, respectively. Furthermore, $\rm LlamaMed\text{-}R^{2}$, which was fine-tuned based on $\rm LlamaMed\text{-}R$, achieved additional gains in both aspects. Compared with $\rm LlamaMed\text{-}R$, $\rm LlamaMed\text{-}R^{2}$ showed improvements of 0.065 in BLEU-1 and 0.062 in ROUGE-L for NLG metrics, and increases of 0.070 in Precision and 0.024 in F-score for CE metrics, confirming that the reflection mechanism effectively enhanced both the linguistic quality and diagnostic accuracy of the generated reports.

\begin{figure*}[!t]
\centering
\includegraphics[width=\linewidth]{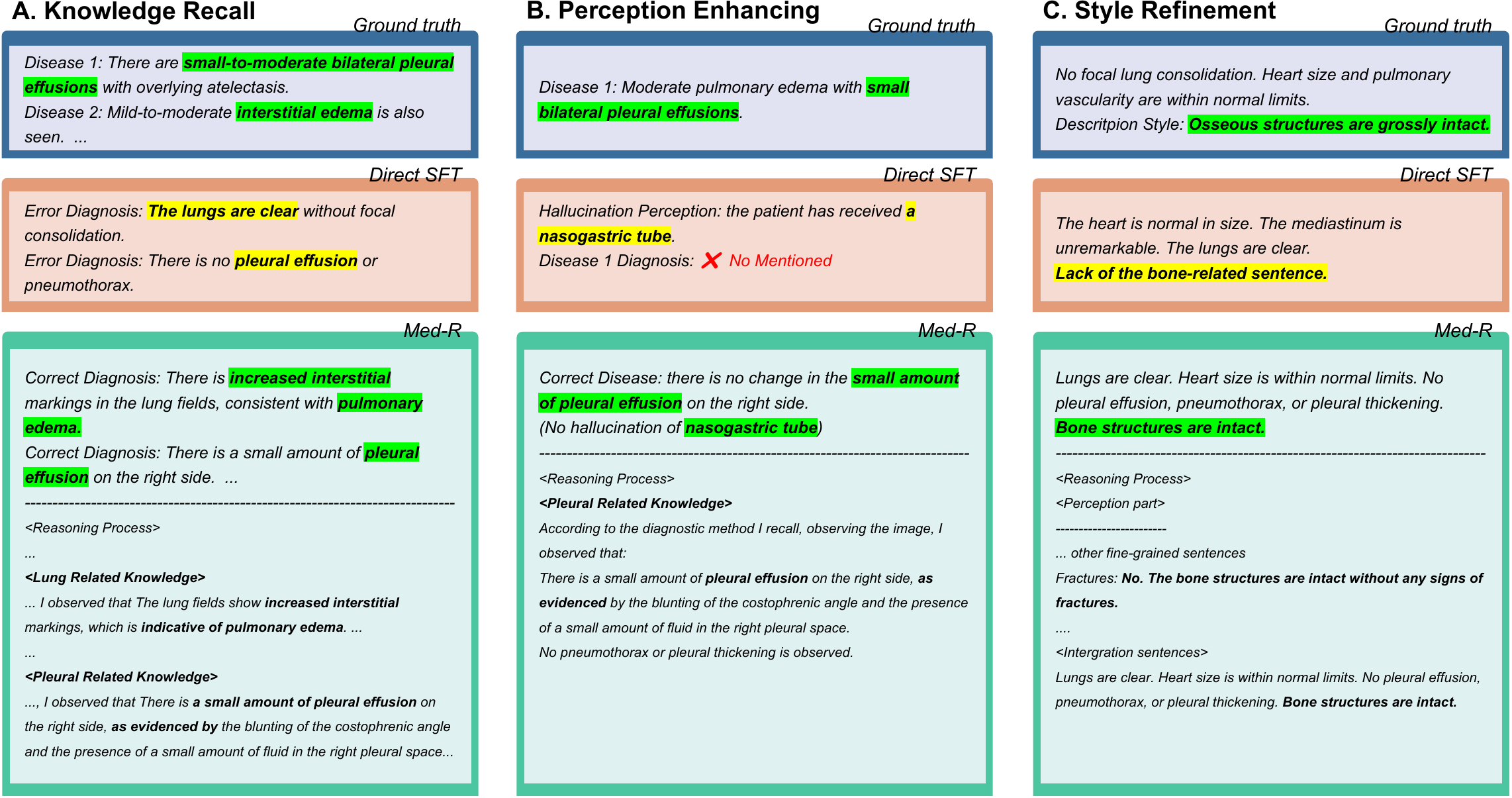}
\caption{Three typical cases generated by LVLMs, which were fine-tuned with direct SFT and our $\rm Med\text{-}R^{2}$. Case A demonstrates the style refinement of the generated report; Case B demonstrates the accurate diagnosis utilizing medical knowledge, and Case C demonstrates the perception-enhancing module that eliminates incorrect recognition. Green highlight indicates ground truth text and correct text generated by our strategy fine-tuned LVLMs, right highlight indicates incorrect text generated by direct SFT LVLMs, and blue highlight indicates related parts in the reasoning process that lead to generating correct text.}
\label{fig:cs1}
\end{figure*}

\subsection{Effectiveness of Complex Reasoning Mechanism}
Fig. \ref{fig:cs1} presents three representative cases demonstrating the advantages of the $\mathrm{Med\text{-}R}$ strategy compared with direct SFT fine-tuning. In Fig. \ref{fig:cs1}A, the ground-truth report includes diagnoses such as 'small-to-moderate bilateral pleural effusions' and 'interstitial edema' for the pleura and lungs, while the direct SFT model incorrectly stated that the lungs were clear and that no pleural effusion was present. After integrating diagnostic knowledge through complex reasoning, the $\mathrm{Med\text{-}R}$ model revised its perception, increasing its alignment with the ground-truth report. It emphasized interstitial markings 'indicative of pulmonary edema' and added 'a small amount of pleural effusion' with supporting evidence, thereby improving disease recognition. In Fig. \ref{fig:cs1}B, the ground truth reports small bilateral pleural effusions, no pneumothorax, and no medical devices, whereas the direct SFT model hallucinated a nasogastric tube. With perception enhancement guided by the perception tree, the $\mathrm{Med\text{-}R}$ model correctly identified a small pleural effusion, retained no pneumothorax, and removed the incorrect device reference, demonstrating reduced false observations and improved perception accuracy. Fig. \ref{fig:cs1}C shows that the ground-truth report concludes with intact bone structures, a detail omitted by the direct SFT model but preserved by $\mathrm{Med\text{-}R}$, indicating that refinement improves both completeness and coherence in clinical reporting.

\begin{figure*}[!t]
\centering
\includegraphics[width=\linewidth]{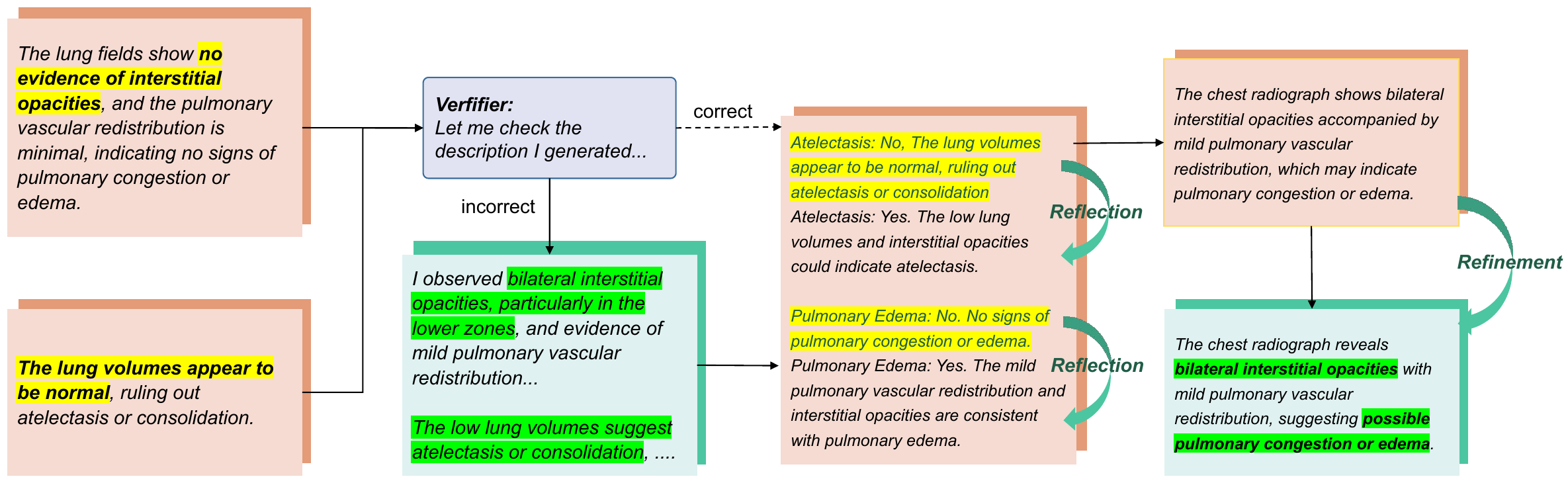}
\caption{A sample patient case illustrating the reflection mechanism applied to the perception and report generation modules. The two identical input text boxes on the left represent the model’s initial generated report before reflection, which the verifier re-evaluates. The red highlights denote erroneous or inconsistent statements, whereas the green highlights indicate the revised descriptions produced through reflection and refinement.}
\label{fig:cs2}
\end{figure*}

\subsection{Influence of Reflection Mechanism}
Fig. \ref{fig:cs2} shows how the reflection mechanism progressively improves both perceptual findings and the final report. Initially, the model incorrectly indicated the absence of abnormalities, e.g. 'no interstitial opacities' and 'normal lung volumes'. After reflection, these errors were corrected, e.g., 'bilateral lower-zone interstitial opacities' and 'low lung volumes suggestive of atelectasis or consolidation', aligning more closely with the ground truth. Correspondingly, the fine-grained diagnostic statements were updated, e.g., earlier 'ruling out of atelectasis or pulmonary edema' was adjusted to 'the low lung volumes and interstitial opacities could indicate atelectasis'. A second reflection then refined wording and clinical tone, adjusting phrasing from a descriptive but less concise form to a clearer and more conventional clinical expression indicating possible pulmonary congestion or edema.

\section{Discussion}
Our results demonstrate that $\mathrm{Med\text{-}R^2}$ fine-tuning strategy achieved the best performance on both MIMIC-CXR and IU-Xray datasets compared to SOTA medical LVLMs (Table \ref{tab:compare}). Compared to direct SFT baselines such as LLaVA-Med and RadFM, our method consistently improved both linguistic quality (NLG metrics) and the reliability of clinical diagnosis (CE metrics). The concurrent gains across these two categories suggest that $\mathrm{Med\text{-}R^2}$ not only generates more coherent reports but also captures effective pathological features as evidence to improve diagnosis accuracy. Furthermore, while MedVLM-R1 introduced explicit reasoning supervision for medical VQA, its transfer to the MRG task yielded only limited CE improvement. We attribute this mainly to the QA-oriented reasoning paradigm that does not align well with the diagnostic process required for report generation. In contrast, $\mathrm{Med\text{-}R^2}$ bridges this gap by coupling perceptual-driven complex reasoning with the reflection mechanism, yielding higher CE scores that reflect both improved rigorous diagnostic reasoning and more effective perception of pathological features. 

As shown in the ablation analysis (Table \ref{tab:ablation}), both $\mathrm{Med\text{-}R}$ and $\mathrm{Med\text{-}R^2}$ consistently outperformed direct SFT across all evaluated backbones. $\mathrm{Med\text{-}R^2}$, by introducing reflection, achieved higher CE precision and F-scores than $\mathrm{Med\text{-}R}$ while maintaining comparable NLG metrics, indicating stronger diagnostic consistency and reliability. $\mathrm{Med\text{-}R}$ enhances diagnostic accuracy by improving perception of effective pathological features and enabling rigorous reasoning grounded in pathological evidence. However, the extended reasoning process occasionally introduced perceptual hallucinations or redundant descriptions. By incorporating a reflection mechanism, $\mathrm{Med\text{-}R^2}$ identified and corrected these errors, resulting in more reliable and accurate diagnostic reports.

The superior performance of $\mathrm{Med\text{-}R^{2}}$ can be attributed to the collaboration of the complex reasoning and reflection mechanisms, which together enhance both perceptual understanding of pathological features and diagnostic reasoning during report generation. The complex reasoning process explicitly decomposes the generation of the medical report into three radiology-specific stages: knowledge recall, pathological perception, and fine-grained report generation. The knowledge recall stage enables the model to retrieve radiological priors to guide effective pathological perception before visual interpretation, allowing it to recognize disease-relevant features with higher precision. The pathological perception stage grounds these recalled concepts into image-level representations, strengthening the model’s ability to localize and describe abnormalities accurately. The fine-grained report generation stage organizes diagnostic reasoning into structured textual outputs, improving linguistic coherence and the logical consistency of clinical conclusions.

Building upon this structured complex reasoning pipeline, the reflection mechanism further refines both the perceptual and linguistic components of the generated reports. By revisiting the perception contents, the model can identify and correct potential misinterpretations or hallucinated findings. Moreover, during the report rewriting phase, it reviews the overall diagnostic narrative to ensure consistency between observations and conclusions, leading to more concise, accurate, and clinically faithful reports. These complementary mechanisms collectively explain the substantial improvements in both NLG and CE metrics, as further validated in the following qualitative analyses.

The qualitative examples in Fig. \ref{fig:cs1} show how the complex reasoning process improves both perception and report generation. In the knowledge recall stage, the model learns to connect medical knowledge with visual features, which helps it focus on disease-related regions and recognize key findings such as pulmonary edema and pleural effusion that are often overlooked by direct SFT baselines. The pathological perception stage further allows the model to describe these findings within their correct anatomical structures, for instance, correctly identifying pleural effusion while avoiding false mentions of nasogastric tubes, making its visual understanding more detailed and accurate. The style refinement stage then improves the organization and clarity of the generated report by presenting observations and impressions in a consistent clinical format, such as adding missing statements on bone integrity. Together, these components establish a structured reasoning framework that allows the model to combine medical knowledge with visual evidence, leading to more accurate and interpretable diagnostic reports.

The reflection process shown in Fig. \ref{fig:cs2} improved complex reasoning quality by enhancing the perception accuracy of pathological features. In the reflection with anatomical perception, two typical errors can be corrected: 1) LVLMs have the awareness of judging a kind of pathological feature, but produce wrong conclusions, such as from 'no evidence of interstitial opacities' to 'bilateral interstitial opacities'. 2) LVLMs observe the key pathological feature, but generate wrong perception contents, such as from 'lung volumes appear to be normal' to 'low lung volumes suggest atelectasis or consolidation'. Furthermore, these cases indicate that the diagnosis accuracy improvement of the generated report is derived from the correction of perception contents. Next, the fine-grained generation of the report was also revised from 'the lung volumes appear to be normal' to 'the low lung volumes and interstitial opacities could indicate atelectasis'. The correction can be broadcast along the reasoning chain and influence the final generated report. At last, some style refinement was applied to the final report to increase the quality of NLG-related metrics.

Although $\mathrm{Med\text{-}R^2}$ demonstrated clear advantages in diagnostic reasoning and report generation, we identified two limitations. Firstly, the extended reasoning and reflection parts increase the token consumption. Despite the increased computation cost, we found it to stay within 1200 tokens and a 30-second generation time, which implies practical usability. Second, the current reflection mechanism lacks fine-grained specificity: it reviews the perception and report as a whole rather than selectively examining pathological features within anatomical structures. Future work will focus on developing feature-level reflection strategies that enable targeted verification of pathological findings and on optimizing the reasoning–reflection pipeline for greater computational efficiency.

\section{Conclusion}
We proposed $\mathrm{Med\text{-}R^{2}}$, a novel two-stage fine-tuning strategy that transforms MRG from a direct mapping task into a structured, perception-driven reasoning process. Unlike the direct SFT strategy, our approach incorporates a complex reasoning pipeline that includes radiology-specific knowledge recall, anatomy-based pathological perception, and fine-grained report generation guided by a clinically informed perception tree. The knowledge recall stage allows the model to retrieve essential radiological priors before visual interpretation, ensuring that the subsequent pathological perception stage grounds these concepts into accurate image-level representations. By explicitly decomposing the diagnostic process, the $\mathrm{Med\text{-}R}$ phase enables LVLMs to focus on disease-relevant features with higher precision, effectively bridging the gap between raw visual data and professional clinical narratives.

To further ensure the reliability of these generated reports, we introduced a reflection mechanism that identifies and corrects potential perceptual hallucinations or logical inconsistencies within the extended reasoning chain. This dual-stage refinement process allows for the correction of erroneous conclusions regarding pathological features while refining the final report's stylistic alignment with clinical conventions. Our extensive experiments on the IU-Xray and MIMIC-CXR datasets demonstrate that $\mathrm{Med\text{-}R^{2}}$ consistently achieves superior performance across both NLG and CE metrics compared to existing SOTA medical LVLMs. Although the extended reasoning process increases token consumption, the significant gains in diagnostic accuracy and interpretive transparency across various LVLMs validate $\mathrm{Med\text{-}R^{2}}$ as an effective solution for automated MRG.

\section{Acknowledgements}
The authors gratefully acknowledge the support from the University of Sydney and the Doubao Medical Team at ByteDance.

\bibliographystyle{abbrv}
\bibliography{references}

\end{document}